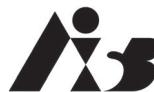
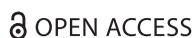
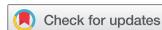

# Interactive natural language acquisition in a multi-modal recurrent neural architecture

Stefan Heinrich and Stefan Wermter

Knowledge Technology Institute, Department of Informatics, Universität Hamburg, Hamburg, Germany

**ABSTRACT**

For the complex human brain that enables us to communicate in natural language, we gathered good understandings of principles underlying language acquisition and processing, knowledge about sociocultural conditions, and insights into activity patterns in the brain. However, we were not yet able to understand the behavioural and mechanistic characteristics for natural language and *how* mechanisms in the brain allow to acquire and process language. In bridging the insights from behavioural psychology and neuroscience, the goal of this paper is to contribute a computational understanding of appropriate characteristics that favour language acquisition. Accordingly, we provide concepts and refinements in cognitive modelling regarding principles and mechanisms in the brain and propose a neurocognitively plausible model for embodied language acquisition from real-world interaction of a humanoid robot with its environment. In particular, the architecture consists of a continuous time recurrent neural network, where parts have different leakage characteristics and thus operate on multiple timescales for every modality and the association of the higher level nodes of all modalities into cell assemblies. The model is capable of learning language production grounded in both, temporal dynamic somatosensation and vision, and features hierarchical concept abstraction, concept decomposition, multi-modal integration, and self-organisation of latent representations.

## 1. Introduction

The human brain is seen as one of the most complex and sophisticated dynamic systems. Humans can build precise instruments and write essays about higher purpose of life because they have reached a state of specialisation and knowledge by externalising information and by interaction with each other. We not only utter short sounds to indicate an intention, but also describe complex procedural activity, share abstract declarative knowledge, and may even completely think in language (Bergen, 2012; Christiansen & Chater, 2016; Feldman, 2006; Håkansson & Westander, 2013). For us, it is extremely easy as well as important to share information about matter, space, and time in complex interactions through natural language. Often it is claimed that language is the cognitive capability that differentiates humans most from other beings in the animal kingdom.

However, humans' natural language processing perhaps is the least well understood cognitive capability. The main reason for this may be the complexity of human language and our inability to observe and study this capability in less complex related species. Another reason is that the neural wiring in the human brain perhaps is not the only component, which is necessary for language to develop. It seems that socio-cultural principles are as well important, and only the inclusion of all factors may allow us to understand language processing. Nevertheless, it is our brain that enables humans to acquire perception capabilities, motor skills, language, and social cognition. The capability for *language acquisition* thus may result from the concurrence of general mechanisms on information processing in the brain's architecture. In particular, in recent studies in neuroscience it was found that the brain indeed includes both hemispheres and all modalities in language processing, and the embodied development of representations might be key in language acquisition (Barsalou, 2008; Glenberg & Gallese, 2012; Hickok & Poeppel, 2007; Huth, de Heer, Griffiths, Theunissen, & Gallant 2016; Pulvermüller & Fadiga, 2010). Furthermore, hierarchical dependencies in connectivity were identified, including different but specific delays in information processing. In linguistic accounts and behavioural studies a number of important principles, such as compositional *and* holistic properties in entities, body-rationality, and social interaction, have been found that might ease – or actually enable – the acquisition of a language competence (Karmiloff & Karmiloff-Smith, 2002; Smith & Kirby, 2012; Smith & Gasser, 2005). In light of the mechanistic conditions of the brain as well as enabling factors of how we learn language *and* other higher cognitive functions, the key objective is to understand the *characteristics* of a brain-inspired *appropriate* neural architecture that *facilitates* language acquisition.

In this paper, we propose[1] a novel embodied multi-modal model for language acquisition to study these important characteristics. As a significant innovation, this model grounds spoken language into the temporal dynamic processing of somatosensory and visual perception and explores a mechanism that abstracts latent representations from these dynamics in a self-organising fashion. Our contribution to knowledge is adding to the understanding of whether connectivity and plasticity attributes of the human brain allow for emergence and development of languages. Results from analytical as well as empirical studies with computer simulations and an interactive humanoid robot will reveal the importance of this self-organisation as well of specific timing in processing speech and multi-modal sensory information.

### 1.1. Previous work on modelling language acquisition and grounding

In the past, researchers have suggested valuable models to explain grounding of language in embodied perception and action, based on neuroscientific data and hypotheses (compare Cangelosi & Schlesinger, 2015; Coradeschi, Loutfi, & Wrede, 2013; Tani, 2014 for an overview). This includes early work on symbol grounding (e.g. Cangelosi, 2010; Cangelosi & Riga, 2006), studies on language evolution and symbol emergence (e.g. Schulz, Glover, Milford, Wyeth, & Wiles, 2011; Steels, Spranger, van Trijp, Höfer, & Hild, 2012), and research sentence comprehension and role filler assignment (e.g. Dominey, Inui, & Hoen, 2009; Dominey & Ramus, 2000). However, due to the tremendous complexity, models are rare which consider the dynamics in full scale and avoid assumptions on predefined word representations (short cutting language processing) or on static or categorically predefined observations (short cutting dynamics in grounding). From studies that approach this complexity, we can adopt important insights.

### 1.1.1. Integrating dynamic vision

Models for grounding in *dynamic vision* are supposed to represent language in the alteration of, for example, perceived objects. Objects can, for example, be altered in terms of changing morphology or motion by self-induced manipulation. Due to complexity, models were often based on a certain decoupling and simplification of the visual stream to achieve a feasible level of coherence in visually perceived features. For example, Yu (2005) developed a model that coupled lexical acquisition with object categorisation. Here, the learning processes of visual categorisation and lexical acquisition were modelled in a close loop. This led to the emergence of the most important associations, but also to the development of links between words and categories and thus to linking similar fillers for a role. The visual features reflect little morphology over time since perception in the visual stream stemmed from unchanging preprocessed shapes in front of a plain background. With changing morphology, Monner and Reggia (2012) modelled grounding of language in visual object properties. This model is designed for a micro-language that stems from a small context-sensitive grammar and includes two input streams for scene and auditory information and an input-output stream for related prompts and responses. In between the input and input-output layer, several layers of long short-term memory blocks are employed to find statistical regularities in the data. This includes the overall meaning of a particular scene in terms of finding a latent symbol system that is inherent in the used grammar and dictionary. Yet, fed in object properties are – in principle – present as given prompts for the desired output responses. This way the emerging symbols in internal memory layers can be determined or shaped by the prompt and response data and are perhaps less latent. Thus it remains unclear how we can relate the emergence of predefined or latent symbols to the problem of grounding natural language in dynamic sensory information to eventually understand how noisy perceived information contributes.

Overall, the studies show that dynamic vision can be integrated as embodied sensation if the dynamics of perception can be reasonably abstracted. For a novel model, however, it is crucial to control complexity in perception to attempt explaining the emerging internal representation.

### 1.1.2. Dynamic multi-modal integration

Integrating multiple modalities into language acquisition is particularly difficult because the linked processes in the brain are extraordinary complex – and in fact – in large parts not yet understood. For this reason, to the best of the authors' knowledge, there is no model available that describes language processing integrated into multi-modal perception with full spatial and temporal resolution for the cortex without making difficult assumptions or explicit limitations. However, frameworks were studied that included temporally dynamic perception that forms the basis for grounding. Marocco, Cangelosi, Fischer, and Belpaeme (2010) defined a controller for a simulated *cognitive universal body* (iCub) robot based on *recurrent neural networks* (RNNs). The iCub's neural architecture was trained to receive linguistic input (bit-strings representing pseudo-words) before the robot started to push an object (ball, cube, or cylinder) and observe the reaction in a sensorimotor way. Experiments showed that the robot was not only able to distinguish between objects via correct "linguistic" tags, but could reproduce a linguistic tag via observing the dynamics without receiving linguistic input and a correct object description. Thus, the authors claimed that the meaning of labels is not associated with a static representation of the object, but with its dynamical properties. Farkaš Malík, and Rebrová (2012) modelled the grounding of words in both, object-directed actions and visual object sensations. In the model, motor sequences were learned by a continuous actor-critic learning that integrated the joint positions with linguistic input (processed in an *echo state network* (ESN)) and a visually perceived position of an object (learned a priori in a *feed-forward network* (FFN)). A specific strength of the approach is that the model, embedded into a simulated iCub, can adapt well to different motor constellations and can generalise to new permutations of actions and objects. However, the action, shape, and colour descriptions (in binary form) are already present in the input of motor and vision networks. Thus, this information is inherently included in the filtered representations that are fed into the model part for a linguistic description. In addition, the linguistic network was designed as a fixed-point classifier that outputs two active neurons per input: one "word" for an object and one for an action. Accordingly, the output is assuming a word representation and omits the sequential order. In a framework for multi-modal integration, Noda, Arie, Suga, and Ogata (2014) suggested integrating visual and sensorimotor features in a deep auto-encoder. The employed time delay neural network can capture features on varying timespan by time-shifts and hence can abstract from higher level features to some degree. In their study, both modalities of features stem from the perception of interactions with some toys and form reasonable complex representations in the sequence of 30 frames. Although language grounding was not pursued, the shared multi-modal representation in the central layer of the network formed an abstraction of perceived scenes with a certain internal structuring and provided certain noise-robustness.

Nevertheless, all in all, we can obtain the insight that forming representation for language can perhaps get facilitated by the shared multi-modal representations and combinations of mechanisms of the brain that filter features on multiple levels.

### 1.2. Paper organisation

This paper is structured as follows: with the related work in mind from the introduction, in Section 2 we will introduce important principles and mechanisms that have been found underlying language acquisition. In Section 3 we will develop a novel computational model by integrating these principles and mechanisms into a general recurrent architecture with aims at neurocognitive plausibility with respect to representation and temporal dynamic processing. We include a complete formalisation to ease re-implementation and will introduce a novel mechanism for unsupervised learning based on gradient descent. Then, in Section 4, we follow up with our evaluation and the analysis. We will specify the scenario for a language learning robot as well as a complete description of the utilised neurocognition-inspired representations for verbal utterances and embodied multi-modal perception. In addition, we report experiments for generalisation and self-organisation. Finally, in Section 5 we will discuss our findings, conclusions, and future prospects.

## 2. Fundamental aspects of language acquisition

Research on language acquisition is approached in different disciplines by complementary methods and research questions. Research in linguistics investigates different aspects of language in general and complexity of formal languages in particular. Ongoing debates about nature versus nurture and symbol grounding led to valuable knowledge of *principles* of learning and *mechanisms* of information fusion in the brain that facilitate language competence (compare Broz, 2014; Cangelosi & Schlesinger, 2015; for a roadmap). Recent research suggests that the well-known principles of statistical frequency and of compositionality in language acquisition are particularly important for forming representation by means of structuring multi-sensory data.

Researchers in different fields related to behavioural psychology study top-down both, the development of language competence in growing humans and the reciprocal effects of interaction with their environment, and have identified important socio-cultural principles. In computational neuroscience, many researchers look bottom-up into the *where* and *when* of language processing and refined the map of activity across the brain for language comprehension and production. New imaging methods allow for much more detailed studies on both, temporal and spatial level, and led to a major paradigm shift in our understanding of language acquisition and underlying mechanisms.

### 2.1. Principles found in developmental psychology

For language acquisition, the first year after birth of a human infant is most crucial. In contrast to other mammals, the infant is not born mobile and matured, but develops capabilities and competencies postnatal (Karmiloff & Karmiloff-Smith, 2002). Development of linguistic competence occurs in parallel – and highly interwoven – with cognitive development of other capabilities such as multi-modal perception, attention, motion control, and reasoning, while the brain matures and wires various regions (Feldman, 2006; Karmiloff & Karmiloff-Smith, 2002). In this process of individual learning the infant undergoes several phases of linguistic comprehension and production competence, ranging from simple phonetic discrimination up to complex narrative skills (Grimm, 2012; Karmiloff & Karmiloff-Smith, 2002).

During this development the infant's cognitive system makes use the following crucial principles among others (Cangelosi & Schlesinger, 2015):

- *Preposition for reference*. The temporally coherent perception of a physical entity in the environment and a describing stream of spoken natural language leads to the association of both (Smith & Yu, 2008).
- *Body-rationality*. Representations, which an infant might form, develop through sensorimotor-level environmental interactions accompanied by goal-directed actions (Piaget, 1954). In addition, the embodiment is suggested as a necessary precondition for building up higher thoughts (Smith & Gasser, 2005).
- *Social cognition*. The development of language is seen only possible by interaction of a child with a caregiver that provides digestible amounts of spoken language (Tomasello, 2003). In particular, mothers provide an age-dependent simplification of grammar and focus on more common words first (Hayes & Ahrens, 1988).

Overall this means that postnatal development of the processes of thought together with an appropriate interaction of a teacher enables acquisition of language.

### 2.2. Mechanistic characteristics found in neuroscience

Based on new imaging methods, several hypotheses have been introduced stating that many cortical areas are involved in language processing. In particular, it was claimed that several pathways between superior temporal gyrus (SFG) and inferior frontal gyrus (IFG) are involved in both language production and comprehension (Friederici, 2012; Hagoort & Levelt, 2009; Hickok & Poeppel, 2007; Huth et al., 2016). These pathways are suggested to include dorsal streams for sensorimotor integration and ventral streams for processing syntax and semantics. An important mechanism found is the activation of conceptual networks that are distributed over sensory areas during processing of words related to body parts (somatosensory areas) or object shapes (visual areas) (Pulvermüller, 2003; Pulvermüller & Fadiga, 2010). Other seemingly important mechanisms found are:

- *Cell assemblies (CAs)*. In higher stages of the spatial or temporal hierarchy, neurons are organised in CAs (Damasio, 1989; Palm, 1990). Those might be distributed over different cortical areas or even across hemispheres and the activation of large and highly distributed CAs can form higher level concepts. Other CAs exist that represent specific semantics like morphemes and lemmas in language processing or are mediators between different levels (Levelt, 2001). The aforementioned conceptual networks can be seen as CAs on word (morpheme) level.
- *Phonological and lexical priming*. The structure of brain connectivity and timing leads to priming, for example, in cohort activation of most relevant sounds or lemmas (Levelt et al., 1991; Marslen-Wilson & Zwitserlood, 1989).
- *Spatial and temporal hierarchical abstraction*. Strongly varying *timescales* take place in the brain. For example in the frontal lobe on caudal–rostral axis, processing of information occurs on much greater timescales from the pre-motor area up to mid-dorsolateral pre-frontal cortex, suggesting that these timings might be relevant for the processing of a plan for motor movement, over sequentialisation, and execution of motor primitives (Badre & D'Esposito, 2009; Badre, Kayser, & D'Esposito, 2010). Similar temporal hierarchies have been found in lower auditory processing (Brosch & Schreiner, 1997; Ulanovsky, Las, Farkas, & Nelken, 2004) and higher vision (Schmolesky et al., 1998; Smith & Kohn, 2008).

Overall this indicates the tight involvement of *general* processes in the brain for reducing and representing complexity in language processing.

### 3. Neurocognitively plausible multi-modal grounding model

Based on aforementioned principles and mechanistic characteristics we can build up a model, which is a neurocognitively plausible *constraint* of a general nonlinear neural architecture. As a starting point we adopt the *continuous time recurrent neural network*

(CTRNN) as a valid abstraction for cortex-level processing (Dayan & Abbott, 2005):

$$\tau \frac{dy_i}{dt} = -y_i + f\left(\sum_{j \in I_{\text{In}}} w_{ij} x_j + b_i + \sum_{k \in I_{\text{Rec}}} w_{ik} y_k\right), \quad (1)$$

where the activity $y$ of a neuron $i$ is derived over time $t$ as an accumulation of previous activity and a function over presynaptic input $x$ (can be sensory input $I_{\text{In}}$, recurrent input $I_{\text{Rec}}$, or both), plastic connections $w$ and a bias $b$. The derivation is governed by a time constant $\tau$ that describes how fast the firing rate approaches the steady-state value. Although we can deduce the CTRNN from the *leaky integrate-and-fire* (LIF) model and thus from a simplification of the Hodgkin–Huxley model from 1952, the network architecture was suggested independently by Hopfield and Tank (1986) as a nonlinear graded-response neural network and by Doya and Yoshizawa (1989) as an adaptive neural oscillator. The CTRNN is thus the most general computational network model as it allows us to define arbitrary input, output, or recurrence characteristics within one (horizontal) layer. Because of the recurrent connections, the network is arbitrarily deep [2] and nonlinear, based on continuous information that are processed over time.

### 3.1. Multiple timescale recurrent neural network

To explore the mechanism of *timescales* as a constraint of the CTRNN, Tani et al. replicated the learning of mammal body motions in an experimental setup along the developmental robotic approach (Nishimoto & Tani, 2009; Tani, Nishimoto, Namikawa, & Ito, 2008; Yamashita & Tani, 2008). These *multiple timescale recurrent neural networks* (MTRNNs) were specified by three layers (called *input-output* (IO) layer, *context-fast* (Cf) layer, and *context-slow* (Cs) layer) with variable timescales and have been trained with a gradient descent method for sequences. The analysis revealed that for a trained network, which could reproduce sequences best (merely indicated by converging to the smallest training error)[3], the patterns in different layers self-organised towards a decomposition of the body movements. The researchers were able to interpret from the neural activity that the Cf layer is always coding for the same short primitive, while the Cs layer patterns are unique per sequence and consist of slow changing values, which function as triggering points for primitives.

#### 3.1.1. MTRNN with context bias

In those original experiments, the researchers were able to train an MTRNN for reasonably diverse and long sequences by initialising the network's neural activity at first time step with specific values of the experimenter's choice (Nishimoto & Tani, 2004; Yamashita & Tani, 2008). These *initial states* were kept for the training of each specific sequence and represented the (nonlinear) association of a constant (starting) value and its dynamic pattern. In later experiments, Nishide et al. adapted and integrated the idea of *parametric bias* (PB) units into the MTRNN (Awano et al., 2010; Nishide et al., 2009). Therein, bias units are part of the Cs layer and parametrise the motion sequence with a certain property (e.g. which tool is used in a certain action), while another initial neural activity is not specified. However, for these bias or *context-controlling* (Csc) units only an initialisation before training is necessary, while the values of these units can self-organise during training. Similar to the *recurrent*

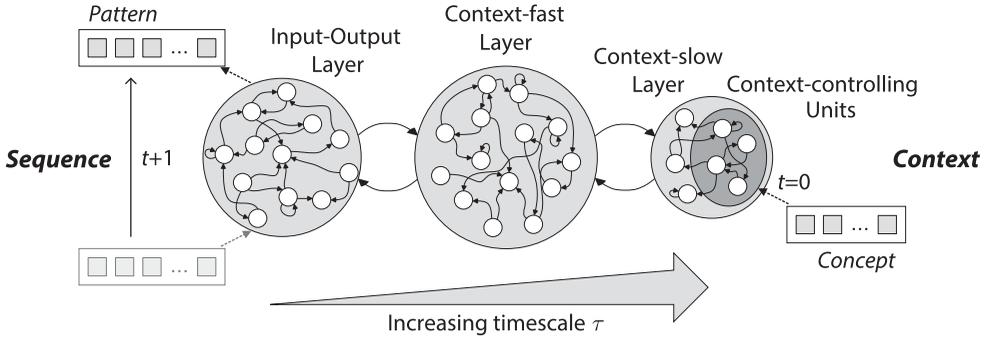

**Figure 1.** The overall MTRNN architecture with exemplary three horizontally parallel layers: *input-output* (IO), *context-fast* (Cf), and *context-slow* (Cs), with increasing timescale $\tau$, where the Cs layer includes some *context-controlling* (Csc) units. While the IO layer processes dynamic patterns over time, the Csc units at first time step ($t = 0$) contain the *context* of the sequence, where a certain concept can trigger the generation of the sequence.

*neural network with parametric bias* (RNNPB), these initial states can be seen as a general context of a sequence. By modulating these internal states, differing other sequences can be generated. Overall, for the conducted experiments on motor primitives, the slow context codes for the general *concept* of a certain body motion.

By combining the characteristics of the various experiments on CTRNNs with multiple timescales and context bias properties (similar to PB but also changing over time), we arrive at a general description of the MTRNN as illustrated in Figure 1. For certain contexts, provided as initial states to some of the neurons with the highest timescale $I_{Csc} \subset I_{Cs}$ (slowest neurons), the network is processing certain sequences over time. The constraints on connectivity and relative timescale setting are inspired by the brain (Badre & D'Esposito, 2009) and have been challenged in developmental robotics studies to confirm a hierarchical compositionality, e.g. in body motion. For further models, we can process dynamic sequences in terms of discretised time steps (e.g. for linguistic processing of smallest graphemic or phonetic units, or visual and sensorimotor processing with a certain sampling rate), but can regard any task as continuous by means of the absolute variability of the timescales.

### 3.1.2. Information processing in the MTRNN

By defining the time constant as a neuron- or unit-dependent variable $\tau_i$ and solving the equation with respect to a time step $t$, we can also describe this special CTRNN in detail[4]: in the MTRNN information is processed continuously with a unit-specific firing rate as a sequence of $T$ discrete time steps. Such a sequence $s \in S$ is represented as a flow of activations of neurons in the IO layer ($i \in I_{IO}$). The input activation $x$ of a neuron $i \in I_{All} = I_{IO} \cup I_{Cf} \cup I_{Cs}$ at time step $t$ is calculated as

$$x_{t,i} = \begin{cases} y_{t-1,i} & \text{iff } t \geq 1 \wedge i \notin I_{IO}, \\ x_{t,i}^* & \text{iff } t \geq 1 \wedge i \in I_{IO,input}, \\ y_{t-1,i}^* & \text{iff } t \geq 1 \wedge i \in I_{IO,output}, \end{cases} \quad (2)$$

where we can either project desired (sensory) input $x^*$ to the IO layer ($I_{IO,input}$) or read out the desired output $y^*$ of the IO layer ($I_{IO,output}$), depending on how the architecture is employed in a task. The input activation for neurons $i \notin I_{\text{IO,input}}$ is initialised with 0 at the beginning of the sequence. The internal state $z$ of a neuron $i$ at time step $t$ is determined by

$$z_{t,i} = \begin{cases} c_{0,i} & \text{iff } t = 0 \wedge i \in I_{\text{Csc}}, \\ \left(1 - \dfrac{1}{\tau_i}\right) z_{t-1,i} + \dfrac{1}{\tau_i} \left(\sum_{j \in I_{\text{All}}} w_{ij} x_{t,j} + b_i\right) & \text{otherwise,} \end{cases} \quad (3)$$

where $c_{0,i}$ is the initial internal state of the Csc units $i \in I_{\text{Csc}} \subset I_{\text{Cs}}$ (at time step 0), $w_{i,j}$ are the weights from $j$th to $i$th neuron, and $b_i$ is the bias of neuron $i$. The output (activation value) $y$ of a neuron $i$ at time step $t$ is defined by an arbitrary differentiable activation function

$$y_{t,i} = f(z_{t,i}) \quad (4)$$

depending on the representation for neurons in IO and on the desired shape of the activation for postsynaptic neurons, e.g. decisive normalisation (softmax) or sigmoidal.

### 3.1.3. Learning in the MTRNN

During learning the MTRNN can be trained with sequences, and self-organises the weights and also the internal state values of the Csc units. The overall method can be a variant of *backpropagation through time* (BPTT), sped up with appropriate measures based on the task characteristics.

For instance, if the MTRNN produces continuous activity (IO) we can modify the input activation with a prorated *teacher forcing* (TF) signal $\alpha \in ]0, 1[$ of the desired output $y^*$ together with the generated output $y$ of the last time step

$$x_{t,i} = \begin{cases} (\alpha) y^*_{t-1,i} + (1-\alpha) y_{t-1,i} & \text{iff } t \geq 1 \wedge i \in I_{\text{IO}}, \\ y_{t-1,i} & \text{iff } t \geq 1 \wedge i \notin I_{\text{IO}}. \end{cases} \quad (5)$$

In the forward pass, an appropriate error function $E$ is accumulating the error between activation values ($y$) and desired activation values ($y^*$) of IO neurons at every time step based on the utilised activation function. In the second step, the partial derivatives of calculated activation ($y$) and desired activation ($y^*$) are derived in a backward pass. In the case of, e.g. a decisive normalisation function (softmax) in IO and a sigmoidal function $f_{\text{sig}}$ in all other layers, we can specify the error on the internal states of all neurons as follows:

$$\frac{\partial E}{\partial z_{t,i}} = \begin{cases} y_{t,i} - y^*_{t,i} + \left(1 - \dfrac{1}{\tau_i}\right) \dfrac{\partial E}{\partial z_{t+1,i}} & \text{iff } i \in I_{\text{IO}}, \\ \displaystyle\sum_{k \in I_{\text{All}}} \dfrac{w_{ki}}{\tau_k} \dfrac{\partial E}{\partial z_{t+1,k}} f'_{\text{sig}}(z_{t,i}) + \left(1 - \dfrac{1}{\tau_i}\right) \dfrac{\partial E}{\partial z_{t+1,i}} & \text{otherwise,} \end{cases} \quad (6)$$

where the gradients are 0 for the time step $T+1$. For the error function $E$ of the decisive normalisation the *Kullback–Leibler divergence* (KLD) is used, where the cross-entropy is generalised to $|I_{\text{IO}}|$ classes (Kullback & Leibler, 1951). Importantly, the error propagated back from future time steps is particularly dependent on the (different) timescales.

Finally, in every epoch $n$ the weights $w$ but also biases $b$ are updated

$$w_{ij}^{(n)} = w_{ij}^{(n-1)} - \eta_{ij}\frac{\partial E}{\partial w_{ij}} = w_{ij} - \eta_{ij}\sum_{t}\frac{1}{\tau_i}\frac{\partial E}{\partial z_{t,i}}x_{t,j}, \quad (7)$$

$$b_i^{(n)} = b_i^{(n-1)} - \beta_i\frac{\partial E}{\partial b_i} = b_i - \beta_i\sum_{t}\frac{1}{\tau_i}\frac{\partial E}{\partial z_{t,i}}, \quad (8)$$

where the partial derivatives for $w$ and $b$ are, respectively, the accumulated sums of weight and bias changes over the whole sequence, and $\eta$ and $\beta$ denote the learning rates for weight and bias changes. To facilitate the application of different methods for speeding up training, we can use individual learning rates for all weights and biases to allow for individual modifications of the weight and bias updates, respectively.

The initial internal states $c_{0,i}$ of the Csc units define the behaviour of the network and are also updated as follows:

$$c_{0,i}^{(n)} = c_{0,i}^{(n-1)} - \zeta_i\frac{\partial E}{\partial c_{0,i}} = c_{0,i} - \zeta_i\frac{1}{\tau_i}\frac{\partial E}{\partial z_{0,i}} \quad \text{iff } i \in I_{\text{Csc}}, \quad (9)$$

where $\zeta_i$ denotes the learning rates for the initial internal state changes.

### 3.1.4. Adaptive learning rates

For speeding up training we employ an adaptation of the *resilient propagation* (RPROP) algorithm that makes use of different individual learning rates $\eta$ and $\beta$ and adapt the learning rates $\zeta$ for the update of the initial internal states $c_{0,i}$ as well (Heinrich, Weber, & Wermter, 2012; Riedmiller & Braun, 1993). In particular, the learning rates $\zeta$ are adapted proportionally to the average of all learning rates $\eta$ over all weights that are connected with unit $i$ and neurons of the same (Cs) and adjacent (Cf) layer

$$\zeta_i \propto \frac{1}{|I_{\text{Cf}}| + |I_{\text{Cs}}|}\sum_{j \in (I_{\text{Cf}} \cup I_{\text{Cs}})} \eta_{ij}. \quad (10)$$

Since the update of the $c_{0,i}$ depends on the same partial derivatives (time step $t=0$) as the weights, we do not need additional parameters in this adaptive mechanism.

### 3.2. Novel unsupervised MTRNN with context abstraction

In the MTRNN with context bias we found that the timescale characteristic is crucial for a hierarchical compositionality of temporal dynamic output sequences. The increasingly slower information processing in the context led to generation of a sequence from an abstract concept. In order to design an architecture that can abstract a context from temporal dynamic input sequences, we can reverse the notion of the context bias and thus reverse the processing from the context to the IO layer. The structure of such a novel MTRNN with context abstraction is visualised in Figure 2. For certain sequential input, provided as a dynamic pattern to the fastest neurons (lowest timescale) $I_{\text{IO}}$, the network is accumulating a common *concept* in the slowest neurons (highest timescale) $I_{Csc} \in I_{Cs}$. Since the timescale characteristics yield a slow adaptation of these so-called Csc units, information in the units will accumulate abstract pattern from the input sequence (filtered by neurons in a potential intermediate layer). The accumulation of information is characterised by a logarithmic skew to the near past and a reach-out to the long past depending on timescale values $\tau_{Cs}$ (and $\tau_{Cf}$).

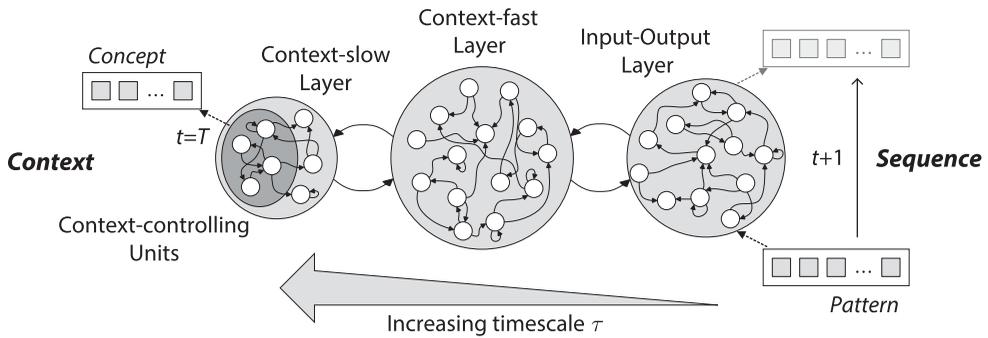

**Figure 2.** The MTRNN with context abstraction architecture providing exemplary three horizontally parallel layers: *context-slow* (Cs), *context-fast* (Cf), and *input-output* (IO), with increasing timescale $\tau$, where the Cs layer includes some *context-controlling* (Csc) units. While the IO layer processes dynamic patterns over time, the Csc units abstract the context of the sequence at *last* time step ($t = T$). The crucial difference to the MTRNN with context bias is an inversion of the direction of procession and an accumulation of abstract context instead of production from a given abstract context.

### 3.2.1. From supervised learning to self-organisation

The MTRNN with context abstraction can be trained in supervised fashion to capture a certain concept from the temporal dynamic pattern. This is directly analogue to fixed-point classification with *Elman recurrent neural networks* (Elman, 1989) or CTRNNs: we can determine the error between a desired temporal static concept pattern and the activity in the Csc units at final time step ($t = T$). With a gradient descent method we can propagate the error backwards through time over the whole temporal dynamic pattern from which the concept was abstracted. However, for an architecture that is supposed to model the processing of a certain cognitive function in the brain, we are also interested in removing the necessity of providing a *desired* target concept a priori. Instead, the representation of the concept should *self-organise* based on regularities latent in the stimuli.

For the MTRNN with PB, this was realised in terms of modifying the Csc units' activity in the initial time step ($t = 0$) backwards by the partial derivatives for weights connecting from those units. Thus the internal states of the initial Csc units self-organised in Csc space towards values that were suited best for generating the sequences of the data set (Hinoshita, Arie, Tani, Okuno, & Ogata, 2011). To foster a similar self-organisation of the Csc units at final time step of the MTRNN with context abstraction, a semi-supervised mechanism is developed that allows us to modify the desired concept pattern based on the derived error.

Since we aim at an abstraction from perception input to the overall concept, the *least mean square* (LMS) error function is modified for the internal state $z$ at time step $t$ of neurons $i \in I_{All} = I_{IO} \cup I_{Cf} \cup I_{Cs}$, introducing a *self-organisation forcing constant* $\psi$ as follows:

$$\frac{\partial E}{\partial z_{t,i}} = \begin{cases} (1-\psi)(y_{t,i} - f(c_{T,i} + b_i))f'_{sig}(z_{t,i}) & \text{iff } i \in I_{Csc} \wedge t = T, \\ \sum_{k \in I_{All}} \frac{w_{ki}}{\tau_k} \frac{\partial E}{\partial z_{t+1,k}} f'(z_{t,i}) + \left(1 - \frac{1}{\tau_i}\right) \frac{\partial E}{\partial z_{t+1,i}} & \text{otherwise,} \end{cases} \quad (11)$$

where $c_{T,i}$ are internal states at the *final* time step $T$ (indicating the last time step of a sequence) of the Csc units $i \in I_{Csc} \subset I_{Cs}$.

The particularly small self-organisation forcing constant allows the final internal states $c_{T,i}$ of the Csc units to adapt upon the data, although they actually serve as a target for shaping the weights of the network. Accordingly, the final internal states $c_{T,i}$ of the Csc units define the abstraction of the input data and are also updated as follows:

$$c_{T,i}^{(n)} = c_{T,i}^{(n-1)} - \psi \zeta_i \frac{\partial E}{\partial c_{T,i}} = c_{T,i}^{(n-1)} - \psi \zeta_i \frac{1}{\tau_i} \frac{\partial E}{\partial z_{T,i}} \quad \text{iff } i \in I_{Csc}, \quad (12)$$

where $\zeta_i$ denotes the learning rates for the final internal state changes. Thereby the learning error $E$ is used in one part ($\psi$) to modify the final internal states and in another part ($1 - \psi$) to modify the weights.

Thus, similarly to the PB units, the final internal states $c_{T,i}$ of the Csc units self-organise during training in conjunction with the weights (and biases) towards the highest entropy. We can observe that the self-organisation forcing constant and the learning rate are dependent, since changing $\zeta$ would also shift the self-organisation – for arbitrary but fixed $\psi$. However, this is an useful mechanism to self-organise towards concepts that are most appropriate with respect to the structure of the data.

### 3.2.2. Preliminarily evaluating the abstracted context

To test in a preliminary experiment how the abstracted concepts form for different sequences using this unsupervised learning mechanism, the architecture was trained for abstracting two contrary cosine waves into context patterns. In particular, for a sequence two cosines waves were presented to two input neurons and discretised to 33 time step. By differently phase-shifting the cosines, four different sequences were prepared. The key aspect of this task is to learn abstract the different phase-shifts in the otherwise identical sequences. In particular because of the ambiguous nature of saddle points, the network cannot simply learn to predict the next time step, but must capture the whole sequence. Processing such a sequence by the MTRNN with context abstraction is supposed to result in a specific pattern of the final Csc units' activity as the abstracted concept.

For determining how those patterns self-organise, the architecture was trained with predefined unchanging patterns (chosen randomly: $\forall i \in I_{Csc}, c_{T,i} \in \mathbb{R}_{[-1.0,1.0]}$) as well as with randomly initialised patterns that adapt during training by means of the varied self-organisation forcing parameter $\psi$. To measure the result of the self-organisation, two distance measures $d_{avg}$ and $d_{rel}$ are used

$$d(c_k, c_l) = \sqrt{\sum_{i \in I_{Csc}} (c_{k,i} - c_{l,i})^2}, \quad (13)$$

$$d_{\text{avg}} = \frac{1}{(|S|-1)\cdot(|S|/2)} \sum_{k=1}^{|S|-1} \sum_{l=k+1}^{|S|} d(c_k, c_l), \quad (14)$$

$$d_{\text{rel}} = \prod_{k=1}^{|S|-1} \prod_{l=k+1}^{|S|} \left(\frac{d(c_k, c_l)}{d_{\text{avg}}}\right)^{1/(|S|-1)\cdot(|S|/2)}, \quad (15)$$

where $|S|$ describes the number of sequences and $c_k = c_{k,T,i}$ denotes the final Csc units of sequence $k$. With $d_{\text{avg}}$, which uses the standard *Lebesgue* $L^2$ or *Euclidean* distance, we can estimate the average distance of all patterns, while with $d_{\text{rel}}$ we can describe the relative difference of distances. For example, in case the distances between all patterns are exactly the same, this measure would yield the best possible result [5] of $d_{\text{rel}} = 1.0$. Comparing both measures for varied settings of $\psi$ provides an insight on how well the internal representation is distributed after self-organisation.

The results for the experiment are presented in Figure 3. From the plots, we can obtain that patterns of the abstracted context show a fair distribution for no self-organisation (the random initialisation) up to especially small values of about $\psi = 1.0 \times 10^{-5}$, a good distribution for values around $\psi = 5.0 \times 10^{-5}$ and a degrading distribution for larger $\psi$. The scatter plots for arbitrary but representative runs in Figure 3(c–f) visualise the resulting patterns for no ($\psi = 0.0$), too small ($\psi = 1.0 \times 10^{-5}$), good ($\psi = 5.0 \times 10^{-5}$), and too large self-organisation forcing ($\psi = 2.0 \times 10^{-4}$). From inspecting the Csc units, we can learn that a "good" value for $\psi$ leads to a marginal self-organisation towards an ideal distribution of the concepts over the Csc space during the training of the weights. Furthermore, a larger $\psi$ is driving a stronger adaptation of the Csc patterns than of the weights, thus leading to a convergence to similar patterns for all sequences.

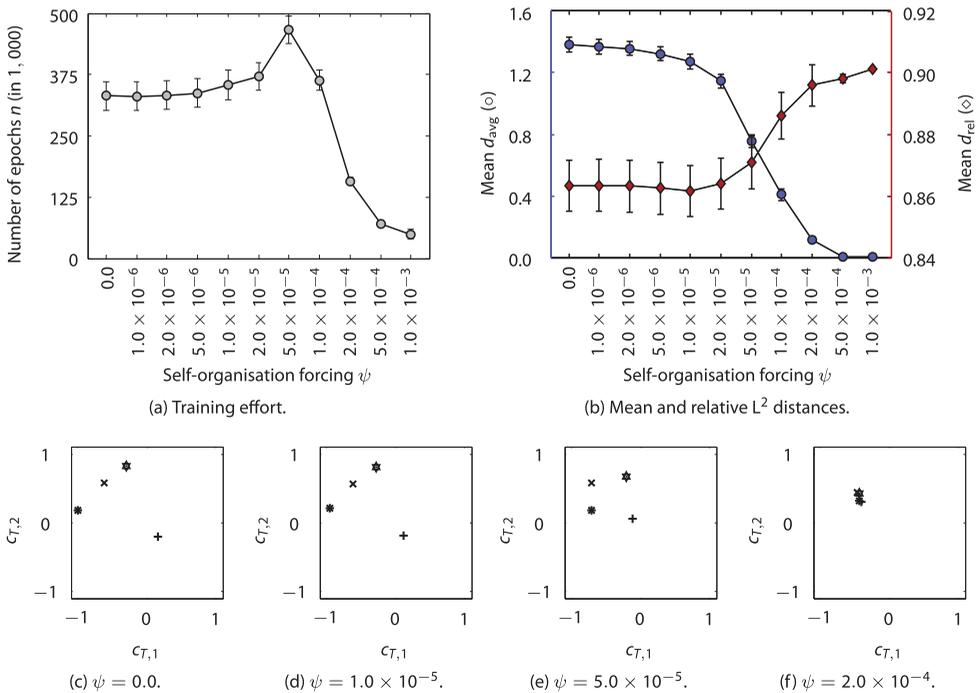

**Figure 3.** Effect of the self-organisation forcing mechanism on the development of distinct concept patterns for different sequences of contrary cosine waves: training effort (a) and mean $d_{\text{avg}}$ and $d_{\text{rel}}$ with standard error bars over varied $\psi$ (b), each over 100 runs; representative developed Csc patterns (c–f) for different sequences for selected parameter settings of no, small, "good", and large self-organisation forcing, respectively.

The task in this preliminary experiment is quite simple, thus a random initialisation within a feasible range of values ($[-1.0, 1.0]$) of the Csc units often already provides a fair representation of the context and allows for convergence to very small error values. However, for larger numbers of sequences, which potentially share some primitives, the random distribution of respective concept abstraction values is unlikely to provide a good distribution, thus self-organisation forcing mechanism can drive the learning.

### 3.3. Previous MTRNN models for language processing

In previous studies, the MTRNN with context bias was tested for modelling language processing due to the mechanism of spatial and temporal hierarchical compositionality. In particular, Hinoshita et al. (2011) utilised the architecture in a model to learn language from continuous input of sentences composed of words and graphemes that stem from a small grammar. For the model, no implicit information is provided on word segmentation and on roles or categories for words. Instead, the input is modelled as streams of spike-like activities on graphemic level. During training, the architecture self-organises to the decomposition of the sentences hierarchically, based on the explicit structure of the inputs and the specific characteristics of some layers. The authors found that the characteristics of information processing on different timescales indeed leads to a hierarchical decomposition of the sentences in a way that certain character orders form words and certain word orders form the sentences. Although the model was reproducing learned symbolic sentences quite well in their experiments, generalisation was not possible to test, because the generation of sentences was initiated by the internal state of the Csc units, which had to be trained individually for every sentence in the model.

Heinrich, Magg, and Wermter (2015) extended this model to process the language embodied in a way that visual input will trigger the model to produce a meaningful verbal utterance that appropriately represents the input. The architecture, called EMBMTRNN model, consists of similar MTRNN layers for the language network, where a verbal utterance is processed as a sequence on phoneme level based on initial activity on an overall concept level. The overall concept is associated with raw feature input over merged shape and colour information of a visually perceived object. Thereby the model incorporates the following hypotheses: (a) speech is processed on a multiple-time resolution and (b) semantic circuits are involved in the processing of language. Experiments revealed that the model can generalise to new situations, e.g. describe an object with a novel combination of shape and colour with the correct corresponding utterance due to the appropriate hierarchical component structure. Yet, in this model the multi-modal complexity of real-world scenarios has not yet been tackled exhaustively. The temporal dynamic nature of visual observations or sensations from another modality was not included and especially not processed on multiple-time resolution.

### 3.4. Novel recurrent neural model with embodied multi-modal integration

Previous models of language processing (compare in Section 3.3) provided insight into the architectural characteristics of language production, grounded in *some* perception. In recent neuroscientific studies, we learned about the importance of *conceptual networks* that are activated in processing speech and that most of the involved processes operate in producing speech as well (compare Borghi, Gianelli, & Scorolli, 2010; Glenberg & Gallese, 2012; Indefrey & Levelt, 2004; Levelt, 2001; Pulvermüller, Garagnani, & Wennekers, 2014). Central findings include that the sensorimotor system is involved in these conceptual networks in general and in action and language comprehension in particular.

For the action comprehension phenomenon, these networks supposedly seem to involve multiple senses. As an example, for actions perceived from visual stimuli, Singer & Sheinberg (2010) found that there is a tight connection between perceiving the form and the motion of an action. A sequence of body poses is perceived as an action if the frames are integrated within 120 ms. Additionally, they found that the visual sequence is represented best as an action if both cues are present, but that in such a case the representation is mostly based on form information. Since *body-rational* motion information is hierarchically processed in proprioception as well, an integration of visual form and somatosensory motion seems more important. These multi-modal contributions – visual and somatosensory – are suggested to be strictly hierarchically organised (compare Friston, 2005; Sporns, Chialvo, Kaiser, & Hilgetag, 2004).

The structure of integration in a conceptual network seems to derive from spatial conditions of the areas on the cortex that have been identified for higher abstraction from the sensory stimuli. These areas, for example , the SFG, but also the IFG, are connected more densely, compared to the sensory regions, but they also show a high interconnectivity with other areas of higher abstraction. From the studies on CAs we deduced that such a particularly dense connectivity, on the one hand, can form general concepts (for example, about a certain situated action) and, on the other hand, may invoke activation first (Pulvermüller et al., 2014).

#### 3.4.1. Model requirements

From these recent findings, hypotheses, and the previous related work, we can adopt that the computational neural model for natural language production should be embedded in an architecture that integrates multiple modalities of contributing perceptual (sensory) information. The perceptual input should also be processed horizontally from sensation encoding over primitive identification (if compositional) up to the conceptual level. Highly interconnected neurons between higher conceptual areas should form CAs and thus share the representations for the made experiences. Importantly the representations should form based on the structure in the perceptual input without a specific target.

In line with the developmental robotics approach (Cangelosi & Schlesinger, 2015), the multi-modal perception should be based on real-world data. Both, the perceptual sensation as well as the auditory[6] production, should be represented neurocognitively plausible. By employing this approach, an embodied and situated agent should be created that acquires a language by interaction with its environment as well as a verbally describing teacher. In this case, the interaction is experienced in terms of the temporal dynamic manipulation of different shaped and coloured objects.

With these requirements, the model implements the principles and mechanistic characteristics described in Section 2.1. Properties of the model supposedly are generalisation despite dynamic embodied perception and disambiguation of inherently focused but limited uni-modal sensation by multi-modal integration. All in all, goals of this model are (a) to refine connectivity characteristics that foster language acquisition and (b) to investigate merged conceptual representation.

### 3.4.2. Multi-modal MTRNNs model

In order to meet the requirements of such a multi-modal model, the following hypotheses are added to the previous EMBMTRNN model (compare Section 3.3) into a novel model named MULTIMTRNNs: (a) somatosensation and visual sensation are processed hierarchically by means of a multiple-time resolution and (b) higher levels of abstractions are encoded in CAs that are distributed over the sensory and motor areas. As a specific refinement of the previous model, the neural circuits for processing the perceptions are modelled each as an MTRNN with context abstraction. The first one, called MTRNN$_s$, processes somatosensation, specifically proprioceptive perception, while the second one, named MTRNN$_v$, processes visual perception. The Csc units of all MTRNNs (within the layers with the highest timescale Cs) are linked as fully connected associator neurons that constitute the CAs for representing the concepts of the information. Based on the abstract concepts the MTRNN with context bias, here called MTRNN$_a$, processes the verbal utterance, again as a sequence on phoneme level. All recurrent neural structures are specifications of a CTRNN to maintain neurocognitive plausibility.

The notation of the IO, Cf, and Cs layers in the novel perception components of the MULTIMTRNNS model, stand for input, fusion, and context of both modalities, somatosensory and vision, respectively. An overview of the architecture is presented in Figure 4. An arising hypothesis for the computational model is that during learning a composition of a general feature emerges, which is invariant to the length of the respective sensory input. A second hypothesis is that features are ambiguous if uni-modal sensations are ambiguous for a number of overall *different* observations, but that the association can provide a distinct representation for the production of a verbal utterance.

### 3.4.3. Information processing, training, and production

For every scene, verbal utterances are presented together with sequences of proprioceptive and visual stimuli of an action sequence. During training of the system, the somatosensory MTRNN$_s$ and the visual MTRNN$_v$ self-organise weights and also the internal states of the Csc units in parallel, for processing of an incoming perception. For the production of utterances, the auditory MTRNN$_a$ self-organises weights and also the internal states of Csc units. The important difference is that the MTRNN$_s$ and the MTRNN$_v$ self-organise towards the *final* internal states of the Csc (end of perception), while the MTRNN$_a$ self-organises towards the *initial* internal states of the Csc (start of utterance). Finally, the activity of the Csc units of all MTRNNs gets associated in the CAs. The output layers of the MTRNN$_a$ are specified by the decisive normalisation, while all other neurons are set up with a sigmoidal function (using a logistic function with $\kappa_h = 0.35795$ for range, and $\kappa_w = 0.92$ for slope as suggested in Heinrich & Wermter, 2014). This particularly includes the neurons in the IO layers of the MTRNN$_s$ and MTRNN$_a$ as well.

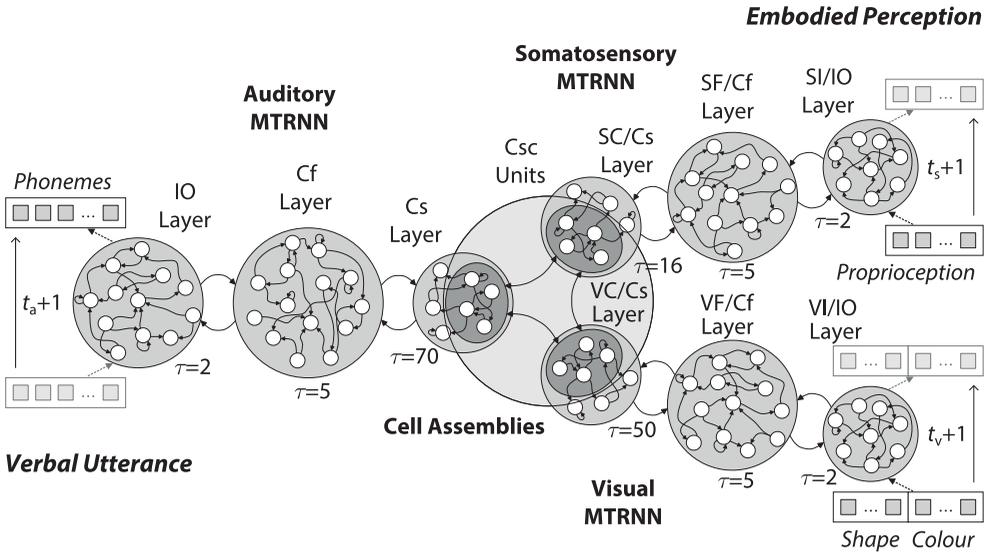

**Figure 4.** Architecture of the multi-modal MTRNN model, consisting of an MTRNN with context bias for auditory, two MTRNNs with context abstraction for somatosensory as well as visual information processing, and CAs for representing and processing the concepts. A sequence of phonemes (utterance) is produced over time, based on sequences of embodied multi-modal perception.

For training of the auditory MTRNN$_a$ the procedure and mechanisms are kept identical to the training in all previous models: the adaptive BPTT variant is utilised by specifying the KLD and the LMS as the respective error functions. The training of the MTRNN$_s$ and MTRNN$_v$ is conducted similarly, but it includes for both the suggested self-organisation forcing mechanism as described in Equation (11) (Section 3.2.1). For these MTRNN with context abstraction, again the error is measured on randomly initialised (desired) activities of the Csc units at the final time step and is used for self-organising both, the weights and the desired internal Csc states. For the CAs, associations between the Csc units of the MTRNN$_s$, MTRNN$_v$, and MTRNN$_a$ are trained with the LMS rule on the activity of the Csc units:

$$\frac{\partial E}{\partial z_i} = (y_i - f_{\text{sig}}(c_{a,0,i}))f'_{\text{sig}}(z_i), \qquad (16)$$

$$z_i = \sum_{j \in I_{s,\text{Csc}}} w_{ij} f_{\text{sig}}(c_{s,T,j}) + \sum_{k \in I_{v,\text{Csc}}} w_{ik} f_{\text{sig}}(c_{v,T,k}) + b_i \quad \forall i \in I_{a,\text{Csc}}, \qquad (17)$$

where $c_{a,0,i}$, $c_{s,T,j}$ and $c_{v,T,k}$ denote the internal states of the Csc units for the MTRNN$_a$, MTRNN$_s$, and MTRNN$_v$, respectively.

With a trained network the generation of novel verbal utterances from proprioception and visual input can be tested. The final Csc values of the MTRNN$_s$ and MTRNN$_v$ are abstracted from the input sequences, respectively, and associated with initial Csc values of the auditory MTRNN$_a$. These values, in turn, initiate the generation of a phoneme sequence. Generating novel utterances from a trained system by presenting new interactions only depends on the calculation time needed for the preprocessing and encoding, and can be done in real time. No additional training is needed.

## 4. Analysis and results

In order to analyse the proposed model's characteristics, we are first of all interested in identifying a parameter setting for the best (relative) generalisation capabilities. Particularly, this enables to analyse the information patterns that emerges for different parts of the architecture. Inspired by infant learning such an analysis will be embedded in a real-world scenario, where a robot learns language from interaction with a teacher and its environment (Cangelosi & Schlesinger, 2015). As a prelude for such an analysis the self-organisation forcing mechanisms need to be inspected further for the impact on the developed internal representation of the abstracted proprioception.

### 4.1. Multi-modal language acquisition scenario

Premised on the principle of social cognition (compare Section 2.1), the scenario is also based in the interaction of a human teacher with a robotic learner to acquire and ground language in embodied and situated experience. For testing the refined model, a *NAO humanoid robot* (NAO) is supposed to learn to describe the manipulation of objects with various characteristics to be able to describe novel manipulation actions with correct novel verbal utterances. Manipulations are to be done by the NAO's effectors and thus to be observed by its motor feedback (proprioception) and visual perception (see Figure 5(a) for an overview). In this study, for the developmental robotics approach, it is particularly important to include the influence of natural variances in interaction, which originate in varying affordances of different objects, but also in unforeseen natural noise.

For a given scene in this scenario, the human teacher guides the robot's arm in an interaction with a coloured object and verbally describes the manipulation action, e.g. "`slide`

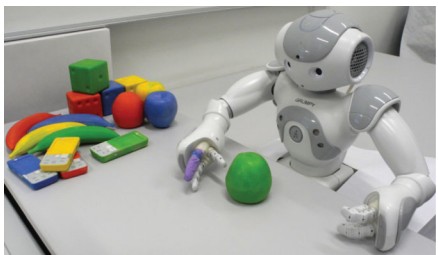

| pull | move the arm behind the object and drag it towards the torso |
| --- | --- |
| push | move the arm in front of the object and push it away from the torso |
| show me | point with the hand to the object |
| slide | move the arm to the right of the object and slide it horizontally to the left |

(a) Scenario overview.   (b) Instructions for manipulation action teaching.

```
Example:
'slide the red apple.'

S   → ACT the COL OBJ.

ACT → pull | push
    | show me | slide

COL → blue | green
    | red | yellow

OBJ → apple | banana
    | dice | phone
```

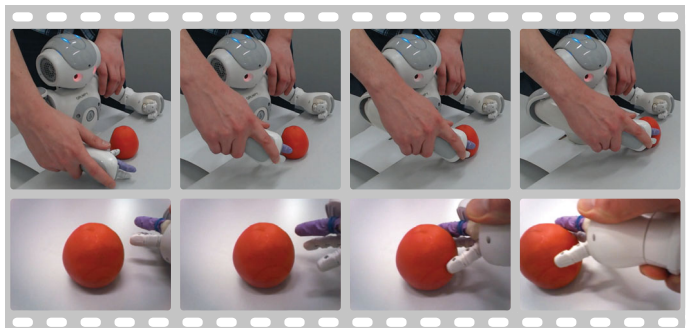

(c) Grammar.   (d) Action teaching over time (bottom: learner's view).

**Figure 5.** Scenario and manipulation action recording for multi-modal language learning scenario.

the red apple". Later, the robot should be able to describe a new interaction composed of motor movements (proprioception) and visual experience that it may have seen before with a verbal utterance, e.g. "show me the yellow apple".

The scenario should be controllable in terms of combinatorial complexity and mechanical feasibility for the robot, but at the same time allow for analysing how the permutation is handled. For this reason, the corpus is limited to a set of verbal utterances, which are generated from the small grammar as summarised in Figure 5(c). For every single object of the same four distinct shapes (apple, banana, phone, or dice) and four colours (blue, green, red, or yellow), four different manipulations are feasible with the arm of the NAO: pull, push, show me, and slide. The grammar is overall unambiguous, meaning that a specific scene can only be described by one specific utterance. Nevertheless, all objects have a similar mass and similar surface conditions (friction). This way the proprioceptive sensation alone is mostly ambiguous for a certain manipulation action on objects with differing colours, but also with different shapes.

In order to collect data for this study, the 64 different possible interactions were recorded 4 times, each with the same verbal utterance and arm-starting position but with slightly varying movements and object placements. This was done by asking different subjects (colleagues from the computer science department) to perform the teaching of such interactions in order to minimise the experimenter's bias (instructions listed in Figure 5(b)).

### 4.1.1. *Neurocognitively plausible encoding*

To encode an utterance into a sequence $s = (p_1, \ldots, p_T)$ of neural activation over time, a phoneme-based adaptation of the encoding scheme suggested by Hinoshita et al. (2011) is used: all verbal utterances for the descriptions are taken from the symbolic grammar, but are transformed into phonetic utterances based on phonemes from the ARPAbet [7] and four additional signs to express pauses and intonations in propositions, exclamations, and questions: $B = \{\text{"AA"}, \ldots, \text{"ZH"}\} \cup \{\text{"SIL"}, \text{"PER"}, \text{"EXM"}, \text{"QUM"}\}$, with size $|B| = 44$. The occurrence of a phoneme $p_k$ is represented by a spike-like neural activity of a specific neuron at relative time step $t_{\text{rel}}$. In addition, some activity is spread backwards in time (rising phase) and forwards in time (falling phase), represented as a Gaußian function $g$ over the interval $[-\omega/2, \ldots, -1, 0, +1, \ldots, \omega/2]$. All activities of spike-like peaks are normalised by a decisive normalisation function for every absolute time step $t$ over the set of input neurons. On the absolute course of time $t$ the peaks mimic priming effects in articulatory phonetic processing. For example, the previous occurrence of the phoneme "P" could be related to the occurrence of the phoneme "AH" leading to an excitation of the respective neuron for "AH", when the neuron for "P" was activated. A sketch of the utterance encoding is shown in Figure 6.

The Gaußian $g$ for $p_k$ is defined by

$$g(p_k, t_{\text{rel}}, i) = \begin{cases} \exp\left(\dfrac{-t_{\text{rel}}^2}{2\sigma^2}\right) & \text{iff } p_k = B_i, \\ 0 & \text{otherwise,} \end{cases} \qquad (18)$$

where $t_{\text{rel}} = 0$ is the mean and the variance $\sigma$ represents the filter sharpness factor. A peak occurs for the neuron $i \in I_{\text{IO}}$ with $|I_{\text{IO}}| = |B|$, if the phoneme $p_k$ is equal to the *i*th phoneme in the phoneme alphabet $B$. From the spike-like activities the internal state $z$ of a neuron $i$

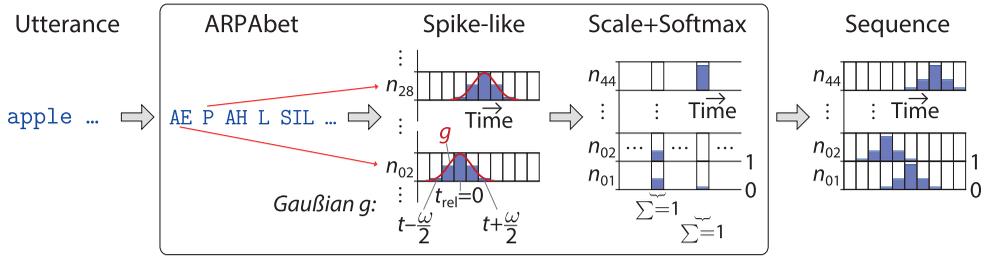

**Figure 6.** Schematic process of utterance encoding. The input is a symbolic sentence, while the output is the neural activity over $|I_{IO}|$ neurons times $T_a$ time steps.

at time step $t$ is determined by

$$z_{t,i} = \begin{cases} \lambda \cdot \max(g(p_{k=1...|s|}, t_{rel} = -\omega/2 \cdots \omega/2, i)) & \text{iff } t = \gamma + k\nu + t_{rel}, \\ 0 & \text{otherwise,} \end{cases} \quad (19)$$

$$\lambda = \ln\left(\frac{0.9}{1.0 - 0.9}(|I_{IO}| - 1)\right), \quad (20)$$

where $\omega$ is the filter width, $\gamma$ is a head margin to put some noise to the start of the sequence, $\nu$ is the interval between two phonemes, and $\lambda$ is a scaling factor for the neuron's activity $y^*$ of maximal values for possibly overlapping spikes. The scaling factor depends on the number of IO neurons and scales the activity to $y^* \in ]0, 0.9]$ for the specified decisive normalisation (softmax) function:

$$y^*_{t,i} = f_{softmax}(z_{t,i}) = \frac{\exp(z_{t,i})}{\sum_{j \in I_{IO}} \exp(z_{t,j})}. \quad (21)$$

For the scenario, the constants are set to $\gamma = 4$, $\omega = 4$, $\sigma^2 = 0.3$, and $\nu = 2$. The ideal neural activation for an encoded sample utterance is visualised in Figure 8(a).

The utterance encoding is neurocognitively plausible because it reflects both, the neural priming mechanism as well as the fluent activation on a spatially distinct phonetic map (Marslen-Wilson & Zwitserlood, 1989; Rauschecker & Tian, 2000). Although research on neural spatial organisation of phoneme coding is in its infancy, there is evidence for an early organisation of the *primary auditory cortex* (A1) and the *superior temporal sulcus* forming a map for speech related and speech unrelated sounds (Chang et al., 2010; Liebenthal, Binder, Spitzer, Possing, & Medler, 2005). The input representation is also in line with an ideal input normalisation to the mean of the activation function, as suggested in LeCun, Bottou, Orr, and Müller (1998).

To gather and encode the proprioception of a corresponding manipulation action, the right arm of the NAO is guided by the human teacher. From this steered arm movement, joint angles of the five joints are directly measured with a sampling rate of 20 *frames per second*. The resulting values are scaled to [0, 1], based on the minimal and maximal joint positions (see Figure 8(b) for an example of the proprioceptive features $F_{pro}$). In a data recording conducted via this scheme, the human teachers are instructed about the four different movements as listed in Figure 5(b). Having an encoding on the joint angle level is neurocognitively plausible because the (human) brain merges information from joint receptors, muscle spindles, and tendon organs into a similar proprioception representation in the S1 area (Gazzaniga, Ivry, & Mangun, 2013). Figure 8(c) shows the encoded proprioception for the exemplary manipulation action.

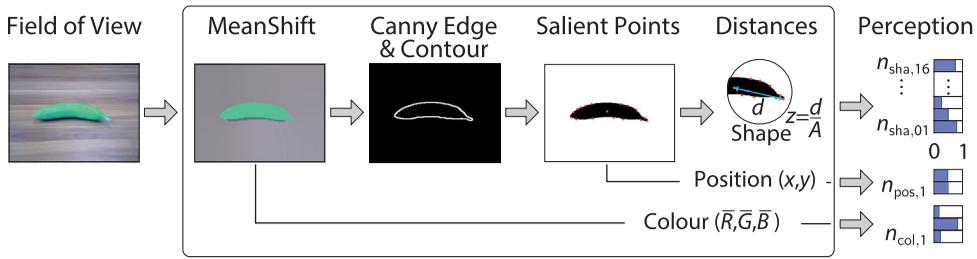

**Figure 7.** Schematic process of visual perception and encoding. The input is a single frame taken by the NAO camera, while the output is the neural activity over $N$ neurons, with $N$ being the sum over shape + colour + positionfeatures.

For the visual perception, we aim at capturing a representation that is neurocognitively plausible but on a level of abstraction of shapes and colours and make use of conventional visual perception methods as shown in Figure 7. At first, the mean shift algorithm is employed for segmentation on an image taken by the robotic learner (Comaniciu & Meer, 2002). The algorithm finds good segmentation parameters by determining modes that describe best the clusters in a transformed 3-D feature space by estimating best matching probability density functions. Secondly, the *Canny edge detection* as well as the OpenCV contour finder are applied for object discrimination (Canny, 1986; Suzuki & Abe, 1985). The first algorithm basically applies a number of filters to find strong edges and their direction, while the second determines a complete contour by finding the best match of contour components. Thirdly, the centre of mass and 16 distances to salient points around the contour are calculated. Here, salient means, for example, the largest or shortest distance between the centre of mass and the contour within intervals of 22.5°. Finally, the distances are scaled by the square root of the object's area and ordered clockwise, starting with the largest. The resulting encoding of 16 values in [0, 1] represents the characteristic shape, which is invariant to scaling and rotation. Encoding of the perceived colour is realised by averaging the three R, G, and B values of the area within the shape. Other colour spaces, e.g. based on only *hue* and *saturation* could be used as well, but they are in this step mainly a technical choice. Additionally, the perceived relative position of the object is encoded by measuring the two values of the centroid coordinate in the field of view to allow for tests on interrelations between multiple objects later.

The resulting encoding is plausible because in the brain is representing visual information in the process of recognising objects similarly by primarily integrating shape and colour features received from the *visual cortex four* (V4) area (Krüger et al., 2013; Orban, 2008). The shape representation codes the discrimination of objects by combining a number of contour fragments described as the curvature-angular position relative to the objects' centre of mass (Pasupathy & Connor, 1999; Yau, Pasupathy, Brincat, & Connor, 2012). The colour representation codes hue and saturation information of the object invariant to luminance changes (Gegenfurtner, 2003; Tanigawa, Lu, & Roe, 2010). For an overview, Figure 8(c) provides two prototypical example results of the perception process, Figure 8(d) provides a sketch of the visual shape perception encoding, and Figure 8(e) shows some of the used objects. The objects have been designed via 3D-print to possess similar masses despite different shapes and comparable colour characteristics across the shapes to provide for robustly and controllably perceivable characteristics.

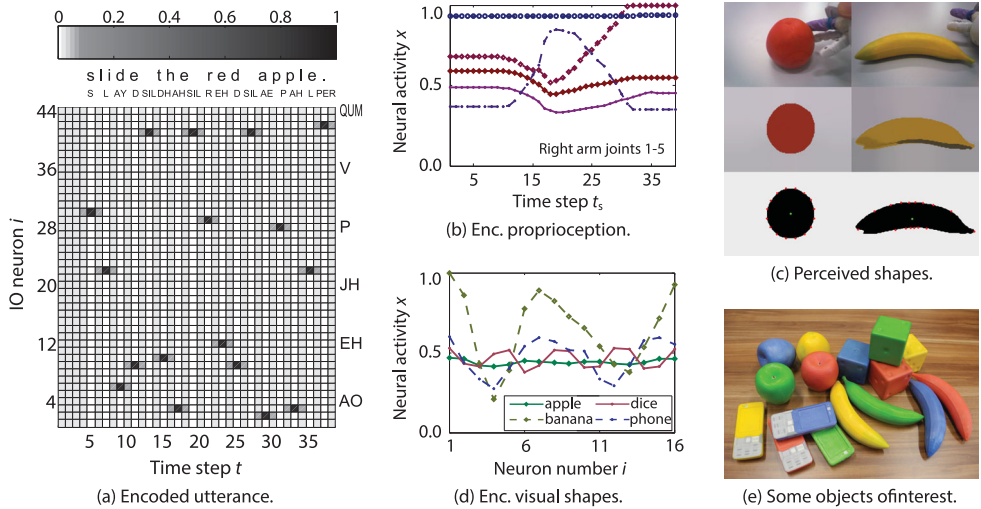

**Figure 8.** Representations in the multi-modal language acquisition scenario.

Capturing motion features also in visual perception is deliberately avoided for several reasons. First of all, from a conceptual perspective, it is desired to keep the visual sensation ambiguous on its own as well as to study the multi-model integration on a conceptual level. Secondly, an agent could experience the movement of an entity in the field of view simply by tracking the said entity with its head or the eyes. This would shift the perception to the somatosensory level and would introduce a redundancy with respect to the arm sensation, which could be difficult to preclude in an analysis.

### 4.2. Experimental setup and evaluation measures

For evaluation, the data were divided 50:50 into training and test sets (all variants of a specific interaction are either in the training or in the test set only) and used to train 10 randomly initialised systems. In addition, this whole process was repeated 10 times as well (10-fold cross-validation) to obtain 100 runs for analysis.

The MTRNNs were parametrised as follows (all parameters are given in Table 1). The auditory MTRNN$_a$ and the visual MTRNN$_v$ were specified in size based on the previous studies for the EMBMTRNN model (Heinrich et al., 2015; Heinrich & Wermter, 2014). The somatosensory MTRNN$_s$ was shaped similarly with $|I_{s,Cf}| = 40$ and $|I_{s,Csc}| = 23$, based on the experience acquired as well as on other work (Yamashita & Tani, 2008). The number of IO neurons in all three MTRNNs were based on the representations for utterances, proprioception, and visual perception and set to 44, 5, and 19, respectively. Also based on previous experience and independent of the data set the number of Csc units were set to $|I_{Csc}| = \lceil |I_{Cs}|/2 \rceil$. All weights were initialised similarly within the interval $[-0.025, 0.025]$, while the initial Csc units (auditory MTRNN$_a$) were randomly taken from interval $[-0.01, 0.01]$ and the final

**Table 1.** Standard meta and training parameter settings for evaluation.

| Parameter | Description | Value | Parameter | Description | Value |
| --- | --- | --- | --- | --- | --- |
| $|I_{a,IO}|$ | Auditory IO neurons | 44 | $\tau_{a,IO}$ | Timescale auditory IO | 2 |
| $|I_{a,Cf}|$ | Auditory Cf neurons | 80 | $\tau_{a,Cf}$ | Timescale auditory Cf | 5 |
| $|I_{a,Cs}|$ | Auditory Cs neurons | 23 | $\tau_{a,Cs}$ | Timescale auditory Cs | 70 |
| $|I_{s,IO}|$ | Somatos. IO neurons | 5 | $\tau_{s,IO}$ | Timescale somatos. IO | 2 |
| $|I_{s,Cf}|$ | Somatos. Cf neurons | 40 | $\tau_{s,Cf}$ | Timescale somatos. Cf | 5 |
| $|I_{s,Cs}|$ | Somatos. Cs neurons | 23 | $\tau_{s,Cs}$ | Timescale somatos. Cs | 50 |
| $|I_{v,IO}|$ | Visual IO neurons | 19 | $\tau_{v,IO}$ | Timescale visual IO | 2 |
| $|I_{v,Cf}|$ | Visual Cf neurons | 40 | $\tau_{v,Cf}$ | Timescale visual Cf | 5 |
| $|I_{v,Cs}|$ | Visual Cs neurons | 23 | $\tau_{v,Cs}$ | Timescale visual Cs | 16 |
| $|I_{a,Csc}|$ | Auditory Csc units | 12 | $\alpha$ | Teacher forcing | 0.1 |
| $|I_{s,Csc}|$ | Somatos. Csc units | 12 | $\eta_{max}$ | Maximal learning rate | 1.0 |
| $|I_{v,Csc}|$ | Visual Csc units | 12 | $\eta_{min}$ | Minimal learning rate | $1.0 \times 10^{-6}$ |
| $C^0_{a,0}$ | Initial Csc values range | ±0.01 | $\xi_+$ | Increasing factor | 1.01 |
| $C^0_{s,T}$ | Init. final Csc r. somatos. | ±1.00 | $\xi_-$ | Decreasing factor | 0.96 |
| $C^0_{v,T}$ | Init. final Csc r. visual | ±1.00 | $\eta^0, \beta^0, \zeta^0$ | Initial learning rates | 0.05 |
| $W^0$ | Initial weights range | ±0.025 | $\psi_v$ | Self-organisation f. visual | $5.0 \times 10^{-5}$ |

Csc units (somatosensory MTRNN$_s$ and visual MTRNN$_v$) from interval $[-1.0, 1.0]$. The learning mechanisms and parameters were identically chosen as in previous studies (Heinrich et al., 2015). Likewise, the timescales for the MTRNN$_a$ and the MTRNN$_v$ were based on the resulting values for the related models ($\tau_{a,IO} = 2$, $\tau_{a,Cf} = 5$, and $\tau_{a,Cs} = 70$) (Heinrich et al., 2015; Hinoshita et al., 2011). A good starting point for the timescale setting of the MTRNN$_s$ were the parameters suggested in original studies ($\tau_{s,IO} = 2$, $\tau_{s,Cf} = 5$, and $\tau_{s,Cs} = 50$) to provide a progressive abstraction (Nishimoto & Tani, 2009; Yamashita & Tani, 2008). For this scenario, the timescales for the somatosensory modality seem not particularly crucial, since the manipulation actions are not strongly dependent on shared motion primitives. A preliminary parameter search (not shown) confirmed these suggestions and revealed good settings for the vision modality in similar ranges ($\tau_{v,IO} = 2$, $\tau_{v,Cf} = 5$, and $\tau_{v,Cs} = 16$).

For the self-organisation forcing parameter of the visual MTRNN$_v$, a parameter exploration was conducted similarly and is excluded here for brevity. This search revealed that the self-organisation is more crucial for this data set, but that a setting of $\psi_v = 5.0 \times 10^{-5}$ again is good .[8]

### 4.3. Generalisation of novel interactions

Based on good parameters for dimensions, timescales, and learning, a variation of the self-organisation forcing parameter $\psi_s$ of the somatosensory MTRNN$_s$ was conducted to test the overall performance of the model. The results of the experiment show that the system is able to generalise well: a high $F_1$-score and a low edit distance (insertion = 1, deletion = 1, substitution = 2) of 0.984 and 0.00364 on the training as well as 0.638 and 0.154 on the test set was determined for the best network. On average over all runs an $F_1$-score and an edit distance of 0.952 and 0.0185 for the training as well as 0.281 and 0.417 for the test have been measured ($q_{F_1-score,mixed} = 0.617$, $q_{edit-dist,mixed} = 0.219$). Note, due to the rigid training scheme there is a high chance that the system had to describe scenes, for which not all aspect (shape, colour, or manipulation action) have been learned before (intended to keep the scenario challenging). For a parameter variation of the self-organisation forcing $\psi_s$ over $\{1, 2, 5\} \cdot 10^{-k}, k \in \{4, 3, 2\}$, all results are provided in Figure 9(a, c). Notably, the best results originated from setting $\psi_s = 5.0 \times 10^{-4}$.

Training is challenging and rarely perfect yet not over-fitted systems were obtained on the training data. Nevertheless, a high precision (small number of false positives) with a lower up to medium recall (not the exact production of desired positives) was observed on the test data. The errors made in production were mostly minor substitution errors (single wrong phonemes) and only rarely word errors.

Using a self-organisation mechanism on the final initial Csc values for the somatosensory and visual MTRNNs caused good abstraction from the perception for the described scenario and the chosen $\psi_s$ and $\psi_v$ values. In this scenario, in fact, the mechanism is very crucial. For both sensory modalities the performance was significantly worse (threshold for $\rho_{t-test} < 0.001$) when using static random values for the final internal states of the Csc units in abstracting the sensation $\psi = 0.0$. In particular for proprioception the rate of successfully described novel scenes nearly doubled when using self-organisation forcing with $\psi_s = 5.0 \times 10^{-4}$ compared to random patterns. Based on the experience acquired in the preliminary test (compare Section 3.2.2), the obvious hypothesis is that the MTRNN$_s$ self-organised a better distribution of the Csc patterns in the Csc space. However, measuring the Csc space by using the $L^2$ distance metrics revealed that the patterns are not spreading out, but rather shrink towards small context values, regardless $\psi_s$ is set too large (see Figure 9(b)): for smaller $\psi_s$ the shrinking develops similar but less strong.

To find an alternative hypothesis, the patterns were inspected again in detail. They showed some regularity for scenes including the same manipulation action. Thus, a good performance might correlate with a self-organisation towards similar patterns for similar manipulations. To quantify this effect, two additional measures are used to describe the difference between patterns for scenes with the same or with different manipulations $M = \{\texttt{pull}, \texttt{push}, \texttt{show me}, \texttt{slide}\}$

$$d_{\text{inter}} = \frac{1}{|M|} \sum_{m_k \in M} d_{\text{avg}}(C_{m_k}), \quad (22)$$

$$d_{\text{intra}} = \frac{1}{(|M| - 1) \cdot (|M|/2)} \sum_{k=1}^{|M|-1} \sum_{l=k+1}^{|M|} d(\text{centroid}(C_{m_k}), \text{centroid}(C_{m_l})), \quad (23)$$

where the inter-cluster distance $d_{\text{inter}}$ is the average of all unweighted pair distances of patterns over the scenes that include the same manipulation (e.g. pull, push, show me, and slide) – subsequently averaged over all manipulations. The intra-cluster distance $d_{\text{intra}}$ provides the mean of all distances of centroids for the clusters $C$ that contain patterns of the same manipulation. The measurements of the inter- and intra-cluster distances over the varied $\psi_s$ are presented in Figure 9(c). The plots are compared on the same absolute scale and show that the inter-distance is decreasing rapidly with increased $\psi_s$, but the intra-distance decreases much slower. At some point, in fact (e.g. for $\psi_s = 5.0 \times 10^{-4}$), the inter-distance is smaller than the intra-distance. This means that the patterns are indeed clustered best for certain $\psi_s$ values before the shrinkage for the Csc patterns is too strong and the distances vanish. In Figure 9(e–g) we can visually confirm this measured clustering on a representative example ("good" in Figure 9(f)).

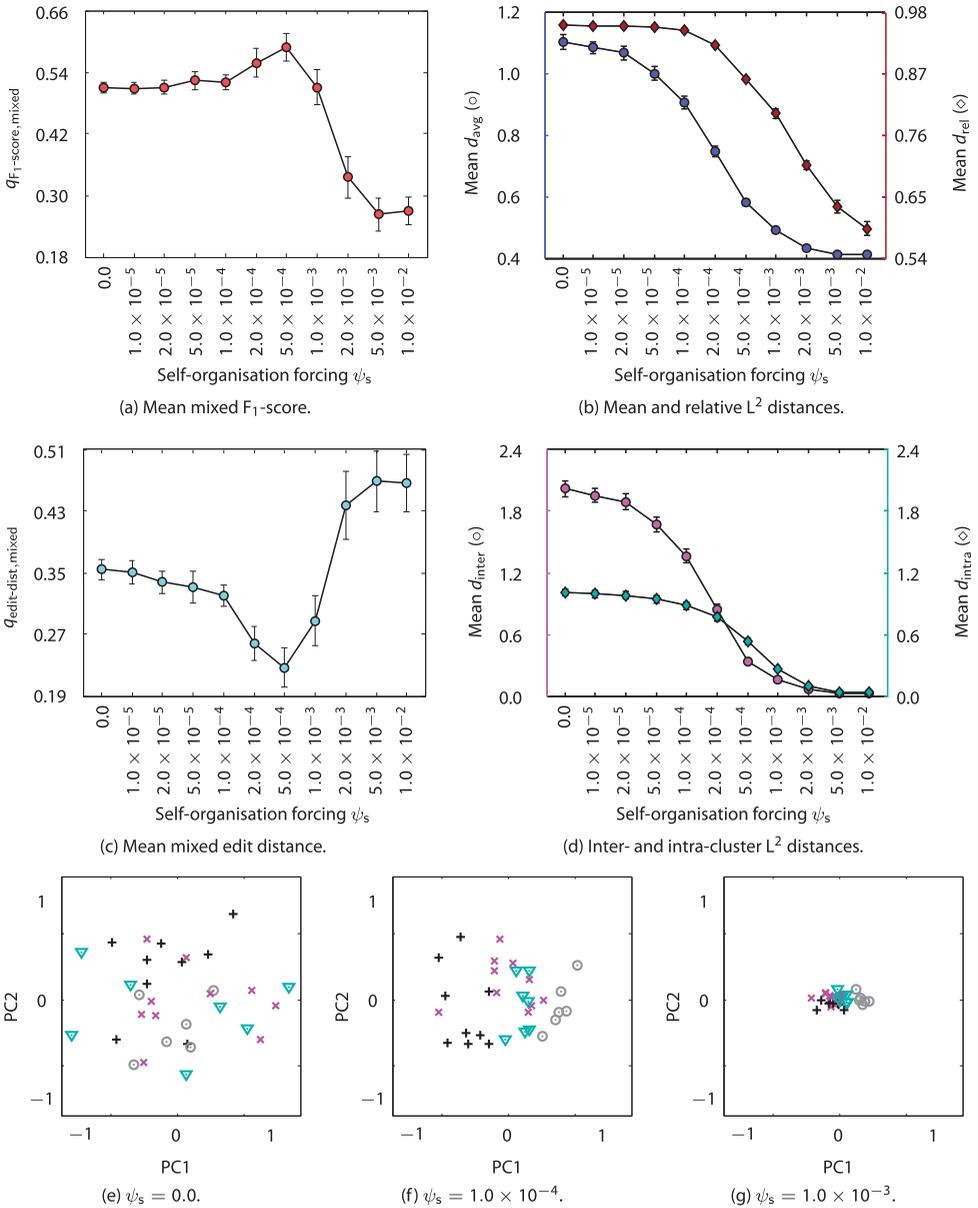

**Figure 9.** Effect of the self-organisation forcing mechanism on the development concept patterns in the MULTIMTRNNS model: mean mixed $F_1$-score (a) and mixed edit distance (b) – "mixed" measures indicate a combination of training and test results with equal weight, mean of average and relative pattern distances (c), and intra- and intra-cluster distances (d) with interval of the standard error, each over 100 runs and over varied $\psi_v$, respectively; representative developed Csc patterns (e–g) reduced from $|I_{Csc}|$ to two dimensions (PC1 and PC2) for selected parameter settings of no, "good", and large self-organisation forcing, respectively. Different words for shapes and colours are shown with different coloured markers (black depicts "position" utterance).

### 4.4. Self-organisation in the CAs

Throughout all tests of the model, diverse patterns of the internal states of the Csc units developed across the modalities. Nonetheless, frequently similar patterns emerged in the respective modality for similar utterances or perceptions. This is particularly the case for the Csc units of the sensory modalities (MTRNN$_s$ and MTRNN$_v$), as shown in the last experiment (where a clustering towards patterns for similar perceptions emerged), but also for Csc units of the auditory production subsequently to the activation within the CAs. During training, the Csc units in the auditory MTRNN$_a$ also self-organised for the presented sequences (utterances). However, within the formation of the CAs by means of the associations, patterns emerged that are able to cover the whole space of scenes in training and test data.

To inspect how these patterns self-organise, we can look into the generated Csc patterns after the whole model is activated by the perception on somatosensory and visual modalities from the training *and* the test data. An example for such Csc activations is presented in Figure 10 for well-converged architectures with a low[9] generalisation rate (a, c, and e) and a high generalisation rate (b, d, and f). The visualisation is provided by reducing the activity of the Csc units to two dimensions using again *principle component analysis* and normalising the values.[10] The results confirm that the patterns form dense and sparse clusters for the visual Csc (the patterns, in fact, overlap each other for different manipulations on the *same* coloured and shaped object). For the somatosensory Csc, the clusters are again reasonable distinct for the same manipulations, although there is a notable mixing between some manipulations on certain objects. For the auditory Csc in case of high generalisation, the patterns are also distinctly clustered. In the example, presented in Figure 10(f), we can discover clustering by colour (prominently on PC2), by manipulation (notably on PC1) and by shape (in between and on lower components). The low generalisation example of Figure 10(e) shows the clusters less clear with more patterns scattered across the PC1 and PC2.

Inspecting the sensory data revealed that visual shape and colour sequences are strikingly similar for different manipulation on the same objects, while the proprioception sequences show some differences for some objects. For example, the slide manipulation on banana-shaped objects was notably different than on the other objects. Apart from that, the proprioception sensation is mostly ambiguous with respect to the specific scene (which object of which shape was manipulated) – which was intended in the scenario design. Thus it seems that in the CAs there is a tendency of restructuring the characteristics (shape, colour, or proprioception), which were overlapping for the single modalities, into a representation where all characteristics are distributed.

### 4.5. Summary

In sum, embedding MTRNNs with context abstraction and an MTRNN with context bias into one coherent architecture allows for a composition of temporal dynamic multi-modal perception into overall concepts and for the decomposition into meaningful sequential actuation, e.g. in terms of a verbal description. From the results, we can deduce that the self-organisation forcing indeed is facilitating the clustering of concepts for similar perceptions by self-organising the space of the internal states of the Csc units upon the structure of the data. Self-organising the patterns in the CAs towards well-distributed clusters highly correlated with the ability to generalise well.

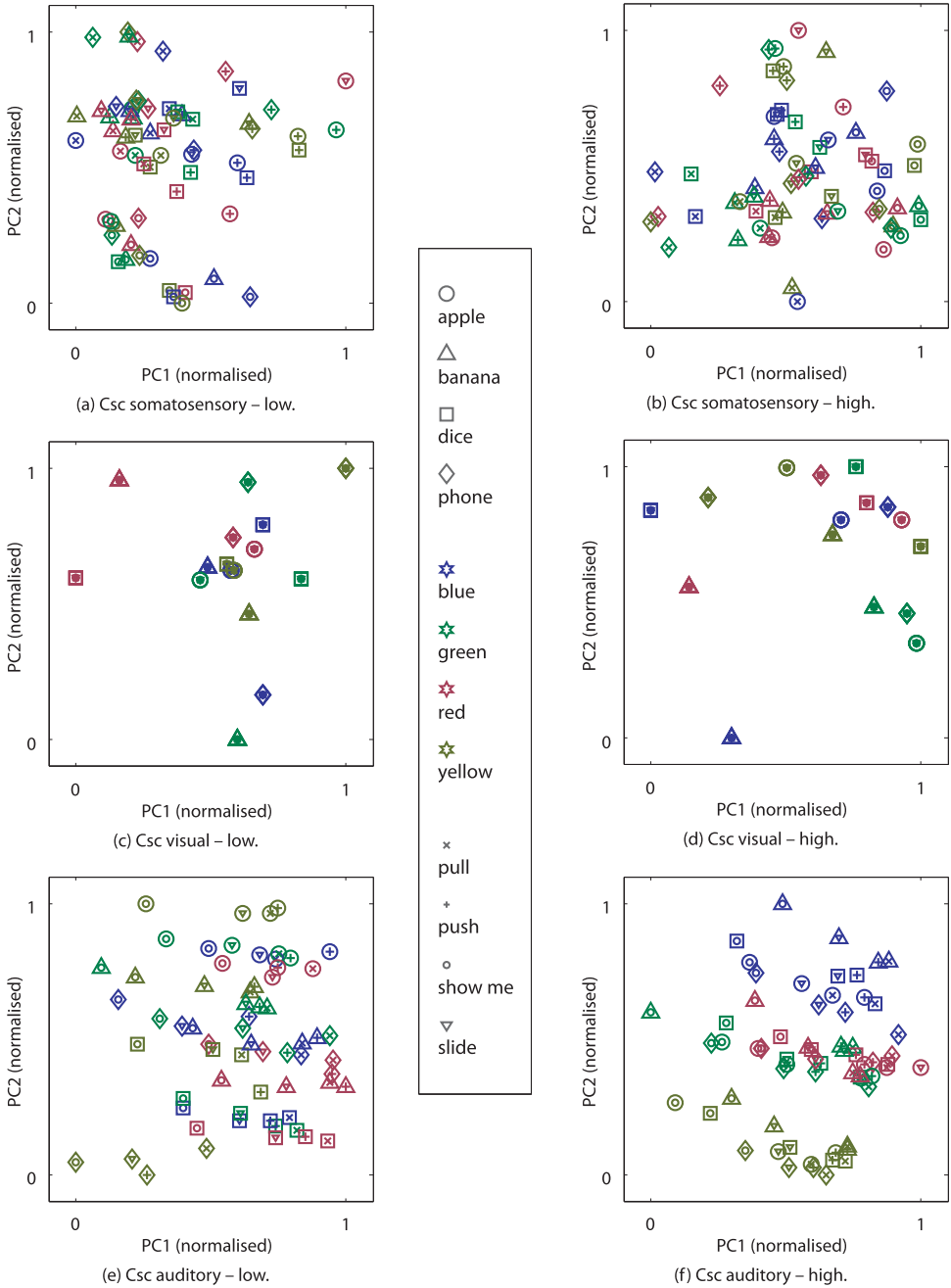

**Figure 10.** Activity in the Csc units after the model has been activated by proprioception and visual perception for the final internal states (somatosensory and visual) and the initial internal states (auditory), reduced from $|I_{Csc}|$ to two dimensions (PC1 and PC2) and normalised, each. Visualisation a, c, e are shown for an representative example for low and b, d, f for high generalisation.

In the novel model, good clustering self-organised for the abstracted context patterns of visual perception and also for somatosensation. For vision, this clustering occurs in particularly dense clusters that are sparsely distributed over the Csc space. For models that generalise well, we found that in the CAs associations emerged that projected the Csc space of the multi-modal sensation (shape, colour, proprioception) into a well-distributed Csc space of auditory production. This distribution self-organised again towards sparsely-distributed dense clusters. Models that are able to successfully describe all training data, but cannot generalise, showed a less well-distributed auditory Csc space.

For the generalisation this means that a well-distributed (sparse) but well-structured (conceptual clusters) auditory Csc space facilitates the grounding of language acquisition into the temporal dynamic features. Such a Csc space allows *modulating* which motor sequence needs to be selected to describe the perception. A good overall abstraction of the respective perceptual features into the CAs thus fosters a correct (good) decomposition into a chain of words and then into phonemes. As a consequence, the CAs fuse but more importantly disambiguate single modalities, which are ambiguous on their own, into an overall coherent representation. Since in the model this happens temporally concurrent, it seems sufficient that different aspects of an observation, just need to co-occur to form a rich but latent overall representation for all modalities.

## 5. Discussion

For the brain, it has been shown that spatial characteristics of neural connectivity and temporal characteristics of neural activation lead to a hierarchical processing of sensation and actuation (compare Section 2). In previous studies researchers have adopted these natural conditions of the cortex in order to model similar hierarchical processing in motor movements and speech production aspects (compare Section 3.1). In particular, these conditions were utilised to constrain CTRNNs with timescales and to integrate a context bias. Such a so-called *MTRNN with context bias* model can decompose an initial context into a sequence of primitives. In this paper, this concept is developed further and reversed to allow for composing a sequence of primitives into an abstracted context. A mechanism is proposed to force an entropy-based self-organisation of such a context, which supposedly serves as a key component of an overall model for grounding language in embodied multi-modal sensation.

### 5.1. Self-organising compositional representations

The self-organisation forcing mechanism provides the development of a latent representation for the respective abstracted context of a sequential perception, without the need of an a priori definition. In the model, the self-organisation forcing parameter is quite sensitive as too small values hinder a self-organisation, while too large values lead to a fast premature convergence of the architecture. A cause for the latter case is that both, the forward activity from small weights as well as a too strong adaptation towards this activity, lead to small errors. Thus, the internal states of the final Csc values are self-organised to match the activity from the network, before the network is self-organised to cover the regularities of the data. This issue could be further approached by using a regularisation for the self-organisation or by using weight initialisations based on the eigenvalue of the weight matrix. For the first option, it would be important to consider methods that are independent of the direction of the gradient. For example, a simple normalisation of the internal states of the final Csc units would only skew the distribution and hence could lead to a convergence towards similar Csc patterns. For the second option, a divergence could occur because the randomly initialised Csc pattern could by chance be all similarly small or similarly large. Utilising weight initialisation and normalisation techniques, used in learning deep FFNs (LeCun, Bengio, & Hinton, 2015), might be interesting but can lead to additional instability during RNN training.

For our model, however, this means that for forming a compositional representation it perhaps is sufficient that the data contain regularities as well as irregularities. It seems that a compositional representation is formed solely by minimising differing activity for similar temporal dynamic patterns (in production and sensation), thus by the entropy of different versus similar patterns. For the concepts of the whole temporal dynamic sequences, this entropy-based descent, which is inherent in the self-organisation forcing mechanism, leads to a restructuring of the concept space to represent similar sequences with similar temporally static concept patterns. All in all, the regularities in the data, which that are also rich in our natural environment (Smith & Gasser, 2005), also seem sufficient for an architecture with different timescales.

### 5.2. Multi-modal context in language acquisition

In the novel model, the density of the formed clusters of certain observations was observed to be closely related to the similarity of the abstracted sequences. This observation is logical since the data for the somatosensory and the visual modalities were not compositional and thus the patterns in the Csc formed as a compression of the temporal dynamic observations. As a consequence, the clustering of sequences is limited by the variability of the sequences, since there is no mapping required to a category within the single modalities. By associating the (clustered) multi-modal sensory representations with the auditory production representations, the CAs form as a direct link of the active patterns. The resulting mappings show a close relation to the action-perception circuits measured in the brain (Pulvermüller et al., 2014): the Csc space is re-organised to form specific conceptual webs for co-occurring multi-modal patterns. Since this effect was not built in but emerged from the entropy-based learning, it seems that the conceptual webs are the obvious consequence of the self-organisation.

Regarding our model, this means that the contexts for the single modalities indeed restructure towards a clustering of similar up to identical patterns for similar perceptions. In this way, the model self-organises towards capturing the features that are different in the otherwise ambiguous sequences. By associating the abstracted temporally static context representations of multiple perception modalities with the speech production modality, concept-level CAs emerge that provide a well-distributed unambiguous context space. Thereby the context space is modulated to produce novel but correct speech productions. With regard to the brain this relates to the finding of synchronous firing between individual neurons, which react to the same stimulus but scaled-up to cortex level (Alho et al., 2014; Engel & Singer, 2001).

Again, both, the uni-modal representations and the associations, self-organise themselves, driven by the regularities in the data. However, the structuring in the single modalities seems less complex and is easier to re-organise. Hence, the hierarchical abstractions seem to operate like a filter on *some* features from the rich perception. Summarising, this means that the multi-modal context is an abstraction for important aspects of the perception on various pathways, to cope with the inherently varying temporal resolutions and information densities of the different modalities.

### 5.3. Conclusion and future work

Overall, in this paper, we present a neurocognitively plausible model for embodied multi-modal language grounding and demonstrate it in a natural interaction of a robotic agent with its environment. The model is an extension of a previous model on embodied grounding, which showed that the spatial and temporal abstraction is an important characteristic for language in the brain (Heinrich et al., 2015), and includes the processing of temporal dynamic somatosensory and visual perception. The characteristics of a neural architecture that facilitating language acquisition that we obtained from the novel model are: (a) shared representations of abstracted multi-modal sensory stimuli and motor actions can integrate novel experience and modulate novel production and (b) self-organisation might occur naturally because of the structure in the sensorimotor data and both, the spatial and temporal nesting that has evolved in the human brain.

Future research must address the demanding training, a scaling-up to larger, and more natural language corpora to cover wider ranges of sensorimotor contingencies. Perhaps we can ease the training by fuzzy characteristics of the neurons, e.g. a stochastic variance in the neurons' firing rate, or the recruitment of new connections without changing the architecture's dimension (LeCun et al., 2015; Murata, Namikawa, Arie, Sugano, & Tani, 2013). In conceptually similar tasks in application, namely sequence to sequence mapping such as video annotation, the machine learning community made tremendous progress recently ( e.g. Donahue et al., 2015; Sutskever, Vinyals, & Le, 2014). Although many utilised architectures employ computational mechanisms that are not neurocognitively plausible, some aspects like pooling, drop-out, and normalisation can be utilised to short-cut parts of the training that are conceptually not crucial for the model, such as pre-processing for visual feature extraction. For scaling-up, the complexity might get reduced by employing the principle of scaffolding in learning a language corpus (Håkansson & Westander, 2013; Rohlfing, Fritsch, Wrede, & Jungmann, 2006; Wrede, Kopp, Rohlfing, Lohse, & Muhl, 2010): words and holo-phrases first, and then more complex utterances without altering the weights from a "word"-layer (e.g. Cf) to the phonetic output. With respect to modelling further phenomena in the brain, it has been suggested that the same conceptual networks may be involved in speech processing, motor action as well as somatosensation (Garagnani & Pulvermüller, 2016; Glenberg & Gallese, 2012). Further refinements can embed hierarchical abstraction and decomposition in utterance comprehension and motor action as well, and test how such a model can replicate an action for verbal descriptions, which were passively learned before or in co-occurrence with the production of an utterance.

With the outcome from our novel model and further refinements, we can design novel neuroscientific experiments on discovering multi-modal integration as well as hierarchical dependencies particularly in language processing and perhaps construct future robotic companions that participate in fascinating discourses.

## Notes

1. Parts of this work have been presented at ICANN 2014 (Heinrich & Wermter, 2014).
2. Compared to non-recurrent FFNs, depth is depending on the arbitrary length of a sequence.
3. The best network during the experiments was shaped by timescale values of 1.0 for IO, 5.0 for Cf, and 70.0 for Cs layers (Yamashita & Tani, 2008).
4. Notation style follows the original description of the MTRNN in Yamashita and Tani (2008).
5. Given the dimensionality of the Csc units is ideal with respect to the number of sequences. However, in cases of a dimensionality that is lower, this value is smaller 1.0. For example, it is not possible to arrange four points in a 2D-plane with equal distance $>0$. In this case, when representing four sequences with two Csc units, we can derive a theoretical optimal $d_{\text{rel}} = 0.9863$. This example can be visualised as having four points in a the 2D-plane arranged optimally on the edges of a square.
6. For consistency with some related literature the speech production, which adopts the involvement of the auditory system, is called auditory production.
7. ARPAbet is a general American English phone set, transcribed in ASCII symbols that was developed in the 1976 Speech Understanding Project by the Advanced Research Projects Agency. Transformation was done using the *Carnegie Mellon University* pronouncing dictionary.
8. Detailed results are omitted, but detailed results for the somatosensory MTRNN$_s$ will be presented within this section.
9. Test set $F_1$-score: low generalisation rate 0.117, high generalisation rate 0.638.
10. The first two components explain the variance in the patterns as follows: low/proprioceptive: 90.75%, low/visual: 52.42%, low/auditory: 83.34%, high/proprioceptive: 97.59%, high/visual: 43.52%, high/auditory: 65.66%.

## Acknowledgements

We would like to thank Sascha Griffiths, Sven Magg, Wolfgang Menzel, and Cornelius Weber for fruitful discussions on model characteristics and experimental design as well as for valuable comments on earlier versions of this manuscript. Also, we want to thank Erik Strahl and Carolin Mönter for important support with the robotic hardware and experimental data collections.

## Disclosure statement

No potential conflict of interest was reported by the authors.

## Funding

We gratefully acknowledge partial support from the German Research Foundation (DFG) under project Crossmodal Learning, TRR-169.

## ORCID

*Stefan Heinrich*  http://orcid.org/0000-0001-9913-3206
*Stefan Wermter*  http://orcid.org/0000-0003-1343-4775